\documentclass{bmvc2k}

\title{Biased Attention: Do Vision Transformers Amplify Gender Bias More than Convolutional Neural Networks?}

\addauthor{Abhishek Mandal}{abhishek.mandal2@mail.dcu.ie}{1}
\addauthor{Susan Leavy}{susan.leavy@ucd.ie}{2}
\addauthor{Suzanne Little}{suzanne.little@dcu.ie}{1}

\addinstitution{
 Insight SFI Research Centre for Data
Analytics \\
School of Computing \\
Dublin City University \\
Ireland
}
\addinstitution{
 Insight SFI Research Centre for Data
Analytics \\
School of Information and
Communication Studies \\
University College Dublin \\
Ireland
}

\runninghead{Mandal, Leavy, and Little}{Biased Attention}

\usepackage{enumitem}
\usepackage{bbm}

\begin{document}

\maketitle

\begin{abstract}
 Deep neural networks used in computer vision have been shown to exhibit many social biases such as gender bias. Vision Transformers (ViTs) have become increasingly popular in computer vision applications, outperforming Convolutional Neural Networks (CNNs) in many tasks such as image classification. However, given that  research on mitigating bias in computer vision has primarily focused on CNNs, it is important to evaluate the effect of a different network architecture on the potential for bias amplification. In this paper we therefore introduce a novel metric to measure bias in architectures, Accuracy Difference. We examine bias amplification when models belonging to these two architectures are used as a part of large multimodal models, evaluating the different image encoders of Contrastive Language Image Pretraining which is an important model used in many generative models such as DALL-E and Stable Diffusion. Our experiments demonstrate that architecture can play a role in amplifying social biases due to the different techniques employed by the models for feature extraction and embedding as well as their different learning properties. This research found that ViTs amplified gender bias to a greater extent than CNNs. The code for this paper is available at: \url{https://github.com/aibhishek/Biased-Attention}
\end{abstract}

\section{Introduction}
\label{sec:intro}
Vision Transformers (ViT), derived from Transformers in Natural Language Processing, have increasingly become important as they outperform Convolutional Neural Networks in many application domains~\cite{kolesnikov2021image, naseer2021intriguing, khan2022transformers}. Unlike Convolutional Neural Networks (CNN), which rely on a sequence of convolution operations extracting information from visual data, ViTs employ Multi-headed Self Attention (MSA) that estimate the relevance of one patch of an image with another~\cite{kolesnikov2021image,khan2022transformers}. This enables ViTs to capture `long-term dependencies' in the data and thus possess a larger receptive field~\cite{khan2022transformers}. Popular computer vision models and their applications have been shown to exhibit a large range of social biases including gender~\cite{buolamwini2018gender,birhane2021multimodal}, racial~\cite{buolamwini2018gender,karkkainen2021fairface}, and geographical biases~\cite{mandal2021dataset,mandal2023gender}. Most of the work done on detecting such biases~\cite{wang2019balanced,serna2021insidebias,zhao-etal-2017-men} and mitigating them~\cite{zietlow2022leveling,wang2020towards,wang2019balanced} are done on CNNs. Although most of the biases originate in the training data~\cite{singh2020female, wang2022revise, mandal2021dataset}, models themselves have been shown to amplify them~\cite{li2015convergent,zhao-etal-2017-men, serna2021insidebias}. Therefore, given the rise in popularity of vision transformers and the lack of previous research on bias detection and mitigation for them, it is crucial to investigate how ViTs handle social biases. 

As the metrics developed for CNNs may not work properly for ViTs~\cite{serna2021insidebias,wang2019balanced}, we introduce a novel bias detection metric: \emph{Accuracy Difference} and adapt the \emph{Image-Image Association Score} developed by~\citet{mandal2023multimodal} to allow comparative analysis between CNNs and ViTs. To detect and study the overall effect of model architecture on gender bias, we analysed the predictions made using models based on these two architectures. We evaluate gender bias with a focus on men and women in this paper, not to reinforce a binary view of gender but with a view to study the effect of bias on model architectures. This paper aims to address the following research questions:
\begin{itemize}[noitemsep,topsep=0pt]
    \item Is gender bias exhibited differently by Convolutional Neural Networks and Vision Transformers?
    \item How can the effect of gender bias in both Convolutional Neural Networks and Vision Transformers be measured?
\end{itemize}

This paper is divided into two parts: The first part measures the effect of gender bias on four sets of CNNs and ViTs using our novel metric and the adapted metric. In the second part, we analyse the zero-shot predictions made by Contrastive Language Image Pretraining (CLIP)~\cite{radford2021learning} using two sets of CNNs and ViTs. We then analyse the results by contrasting the differences between these two model architectures. Additionally, for our metrics, we created an occupation-based visual dataset by crawling images from the Internet.  

\section{Background and Related Work}
\citet{parkvision} studied the various differences between CNNs and ViTs and found two key differences: the shallower learning profile for ViTs leading to better generalisation when trained on large datasets and Multiheaded Self Attention (MSA) being high pass filters and Convolutions being low pass filters. MSA enables ViTs to model full image contextual information and, coupled with the flatter loss landscape, enables ViTs to attain better generalisation and model long-range contextual information than CNNs when trained on large datasets~\cite{khan2022transformers}. The absence of inductive priors (which are present in CNNs) allows ViTs to attain global attention and better learn contextual cues~\cite{girdhar2019video}.  

\textbf{Measuring bias in deep neural networks: } Several metrics such as Image Embeddings Association Test \cite{sirotkin2022study}, model leakage and bias amplification \cite{wang2019balanced}, and InsideBias \cite{serna2021insidebias} have been proposed to detect and measure gender bias in vision models. However, they have been mainly developed for and tested on Convolutional Neural Networks. With the increasing adoption of Vision Transformers, it is important to develop similar metrics for ViTs. 

\textbf{Image-Image Association Score (IIAS)} developed by \citet{mandal2023multimodal} measures stereotypical associations in vision models. It is derived from the Word Embeddings Association Test in Natural Language Processing, which is itself based on the highly popular Implicit Association Test. It estimates human-like biases in vision models by measuring the association between two sets of concepts: two attributes and a target in the model's embeddings. The attributes in the case of gender can be man and woman, and the target can be a real-world concept like occupation. Thus, if a particular occupation (e.g. CEO) is closer to man than woman, in a model's embedding space, then the model is biased.

\textbf{Contrastive Language Image Pretraining (CLIP)} is a large multimodal model developed by OpenAI, trained on 300 million image-text pairs crawled from the Internet~\cite{radford2021learning}. It connects images with text and is trained using contrastive loss and is used in other popular generative models such as DALL-E and Stable Diffusion~\cite{ramesh2022hierarchical,rombach2022high}. CLIP uses a text encoder and an image encoder, with the option of CNNs (ResNet 50,50x4, and 101) and ViTs (ViT B/16 and B/32) being provided. This enables us to study the multimodal effect of bias in these two architectures from a multimodal perspective. Although CLIP has been shown to exhibit social biases~\cite{radford2021learning, wolfe2022contrastive, wolfe2022evidence}, the effect of image encoder architecture on bias is yet to be studied. 

\section{Measuring Bias}

\subsection{Accuracy Difference}
For a multiclass, class-balanced visual dataset $\mathcal{D}$ containing instances $(X_i , Y_i , g_i)$, where $X_i$ is an image having class label $Y_i$, and a protected attribute $g_i$ denoting gender, where $g_i \in \{m,w\}, (m: men, w: women)$.  
Let $\mathcal{D}_{balanced} \subset \mathcal{D};$  $f(g_i (m=w))$, be a dataset containing instances with protected attributes such as gender. The dataset is class balanced as well as gender-balanced, meaning all instances have an equal gender ratio. Let $\mathcal{D}_{imbalanced} \subset \mathcal{D};$ $f(g_i (m>w \vee m<w))$, be a dataset which is class-balanced but gender imbalanced. Let $\mathcal{D}_{test} \subset \mathcal{D}$ be a class and gender-balanced dataset. 
The generalisation error (misclassification rate) of a classifier trained on $\mathcal{D}$ and tested on $\mathcal{D}_{test}$ can be estimated as:
    
\[
E = \frac{1}{N} \sum_{i=1}^{N} \mathbbm{1} (y_i \neq \hat{y_{i}}) \qquad \cdots eq (1)
\]
where $\mathbbm{1}(.)$ is the indicator function, $N$ is the number of samples in the dataset, and $\hat{y_{i}}$ is the predicted class label. The generalisation error (misclassification rate) can also be given as:
\[
E = bias + variance + unavoidable~error\qquad \cdots eq (2)
\]
If we neglect the unavoidable error and express bias and variance in terms of $g_i$, then $g_i$ can be used as a proxy for $E$. As the accuracy of the classifier on the $\mathcal{D}_{test}$ can be expressed as $1-E$, then from $eq (1) $ and $eq (2)$, accuracy can be used as a proxy for bias $g_i$. 
Let image classifiers $M_{unbiased}$ be trained on $\mathcal{D}_{balanced}$  and $M_{biased}$ be trained on $\mathcal{D}_{imbalanced}$ having an accuracy of $A_{biased}$ and $A_{unbiased}$ on $\mathcal{D}_{test}$ respectively. 

Then we define accuracy difference ($\Delta$) as:
\[
\Delta = \lvert A_{unbiased} - A_{biased} \lvert  \qquad \cdots eq (3)
\]
If the effect of gender bias on a classifier is minimal, then $M_{biased}$ will perform very similarly to $M_{unbiased}$ on the gender-balanced $\mathcal{D}_{test}$ and $\Delta$ will be very small. However, if the effect of gender bias on the classifier is significant, then the performances of the models will differ and $\Delta$ will be high. Higher the value of $\Delta$, more the effect of bias.

\subsection{Image-Image Association Score (IIAS)}
The authors of IIAS~\cite{mandal2023multimodal} used CLIP embeddings to calculate IIAS. We adapted the metric by replacing the CLIP embeddings with the image features extracted by the classifier model. In the case of CNNs, it was the output of the final pre-fully connected layer and in the case of ViTs, the final pre-MLP layer. We then used cosine distance to measure similarity. For two images \(I_1 \) and \(I_2 \), with extracted features \(\nu_1 \) and \(\nu_2 \)  respectively, we calculate image similarity as:  
\[
 sim(I_1,I_2)  = \frac{\nu_1 \cdot \nu_2}{||\nu_1||_2\cdot||\nu_2||_2} \quad \ldots eq(4)
\]
\[
 sim(I_1,I_2) \in [0,1]
\]
Then we calculate IIAS in the same way as the authors. Let $A$ and $B$ be two sets of images containing images of men and women, respectively called gender attributes. Let $W$ be a set of images containing images corresponding to a real-world concept such as occupation, called target. Then the Image-Image Association Score, IIAS, is given by:
\[II_{AS} = mean_{w \in W}s(w,A,B) \quad \ldots eq(5)
\]
where,
\[s(w,A,B) = mean_{a\in A}sim(\vec w, \vec a) - mean_{b\in B}sim(\vec w, \vec b) \qquad [from \quad eq (4)]
\]
\[
IIAS \in [-1,1]
\]
If IIAS is positive, then the target is closer to men showing a male bias and if IIAS is negative, then the target is closer to women, showing a female bias. The numeric value indicates the magnitude of the bias.

\section{Experiment}
The experiments are divided into two parts. In the first part, we measure the effect of gender bias on eight sets of image classifiers belonging to CNNs and ViTs, using Accuracy Difference and IIAS. In the second part, we analyse the zero-shot predictions of CLIP using four different image encoders belonging to CNNs and ViTs.

\subsection{Bias Analytics using Image Classifiers}
We selected four CNN models: VGG16, ResNet152, Inceptionv3, and Xception, and four ViT models: ViT B/16, B/32, L/16, and L/32. All the models were pre-trained on the Imagenet dataset. We used the feature-extracting layers of the models and added customised dense layers to all the models. Then, the models were fine-tuned and tested on our custom dataset containing about 10k images. In order to ensure controlled variables, we limited our study to simpler models such as the original ViTs and older CNNs. This allowed us to isolate the bias comparison solely to the architecture and not have any influence from complex additions.

\subsubsection{The Dataset}
We created a custom visual dataset to measure gender bias by crawling images using Google Search using the Selenium library\footnote{https://www.selenium.dev/} for occupation-related query terms `CEO', `Engineer', `Nurse', and `School Teacher'. The occupation categories `CEO' and `Engineer' are traditionally male-dominated and `Nurse' and `School Teacher' are female-dominated~\cite{wang2022revise,singh2020female,mandal2021dataset}. Two sets of training data were created: gender-balanced and imbalanced. In the balanced dataset, all categories have a 50:50 split of images of men and women. In the imbalanced dataset, the gender ratio of the classes was split in a male:female ratio of 9:1 for `CEO' and `Engineer' and 1:9 for `Nurse' and `School Teacher', as per existing workforce bias. The queried images did show gender bias as per previous research~\cite{mandal2021dataset,wang2022revise} and the gender ratio was adjusted in order to achieve uniformity. The test dataset was also gender balanced. The image filtering to achieve the necessary gender ratios was done manually.  The train dataset consists of 7,200 images: 3,600 images for balanced and imbalanced datasets with each containing 900 images for each category. The test dataset consists of 1,200 images: 300 images for each category with 150 images for each gender. The validation sets for both the biased and unbiased training were split from the balanced and imbalanced datasets manually, keeping the gender ratios intact. A separate dataset containing images of men and women was queried using the terms `man' and `woman' for the IIAS assessment.

\subsubsection{Measuring Accuracy Difference}
The models were partially retrained (fine-tuned) on the balanced and imbalanced datasets, creating a total of 80 models: (4 CNNs \& 4 ViTs) x 2 (biased \& unbiased) x 5 iterations. The training methodology for the CNNs is as follows. First, the feature-extracting layers were frozen and the custom dense layers warmed up for 50 epochs. Then the last two convolution blocks were unfrozen and the model was trained for a further 50 epochs with a smaller learning rate and with early stopping parameters with patience set to 10 iterations. For the ViTs, first the feature extracting layers were kept frozen and the models trained for 100 epochs with early stopping parameters with patience set to 10 iterations. Then the entire model was unfrozen and trained for 50 epochs with a very small learning rate with early stopping patience set to 5. The Accuracy Difference was calculated for all the models as explained in section 3.1 and as per eq(3). 

\subsubsection{Measuring IIAS}
\begin{figure}[ht]
\centering
\includegraphics[scale=0.4]{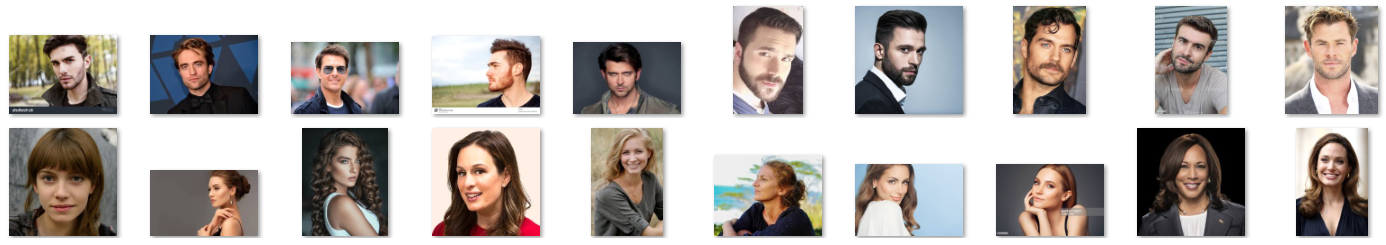}
\caption{Gender attributes - Men (top) and Women}
\label{fig:fig_1}
\end{figure}

The fine-tuned biased and unbiased models (from the previous experiment; section 4.1.2) were saved and their classification layers were removed for this part and the models were used as feature extractors on two sets of target images. The first set is the test dataset used for the previous part and for the second set, we blacked out (masked) the faces in the images as the most important feature for determining gender. Two sets of five images of men and women each were used for each part (masked and unmasked) as targets (Table~\ref{tab:table_1}). Ten images of men and women each were used as gender attributes (Figure~\ref{fig:fig_1}). Then, the biased and unbiased model feature extracting layers were used to calculate IIAS as per eq (5). The experiment was repeated five times and the images for the attributes and the targets were chosen randomly without repeating. It is important to note that only the last layers of the CNN based feature extractors were retrained on our dataset, but as the training data for all the models are the same, it gives us an estimate of how bias is handled differently by the different model families.

\subsection{Bias Analytics using CLIP}
To further understand the effect of gender bias on model architecture, four different types of CLIP image encoders were used: CNNs ResNet 50 and 50x4 and ViTs ViT B/16 and B/32. A list of 100 occupation terms was created based on official lists and CLIP's zero-shot predictions used to predict labels for images of men and women (full list of terms is provided in Appendix A). The image dataset is the same as that used for attributes in the IIAS experiment. The top predictions for men and women were then analysed to study the differences in the effect of gender bias on CNNs and ViTs.

\begin{table}[]
\centering
\begin{tabular}{|l|l|l|}
\hline
                                                          & \textbf{Masked} & \textbf{Unmasked} \\ \hline
CEO                                                       & \includegraphics[width=0.2\textwidth]{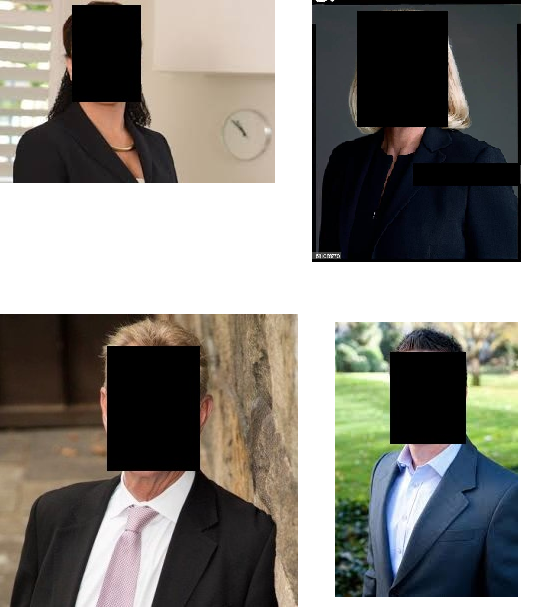}          & \includegraphics[width=0.2\textwidth]{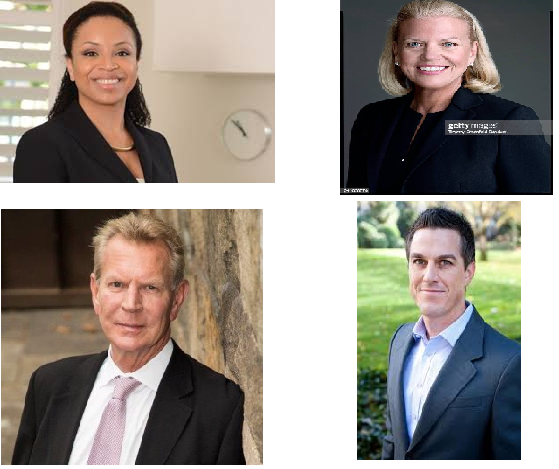}            \\ \hline
Engineer                                                  & \includegraphics[width=0.2\textwidth]{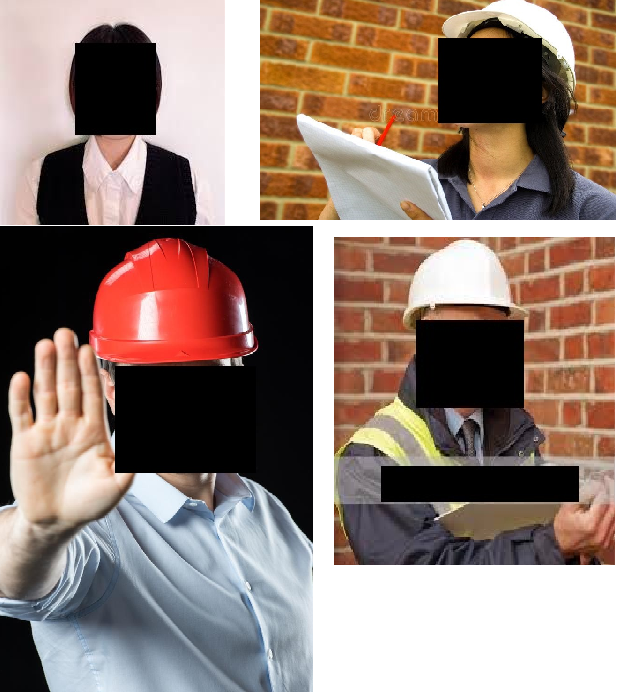}          & \includegraphics[width=0.2\textwidth]{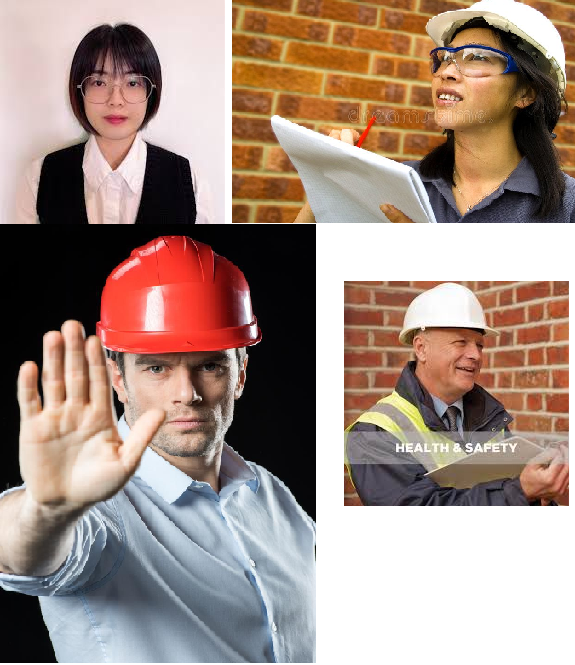}          \\ \hline
Nurse                                                     & \includegraphics[width=0.2\textwidth]{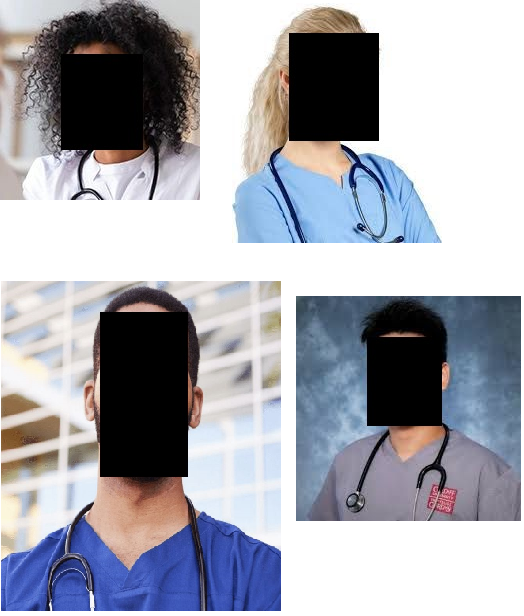}          & \includegraphics[width=0.2\textwidth]{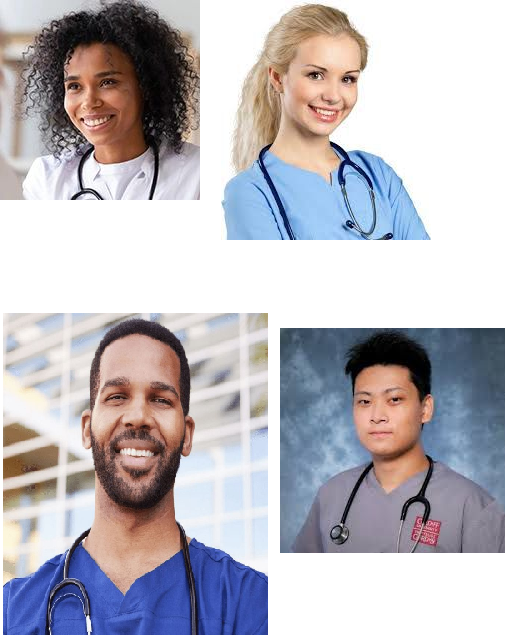}           \\ \hline
\begin{tabular}[c]{@{}l@{}}School \\ Teacher\end{tabular} & \includegraphics[width=0.2\textwidth]{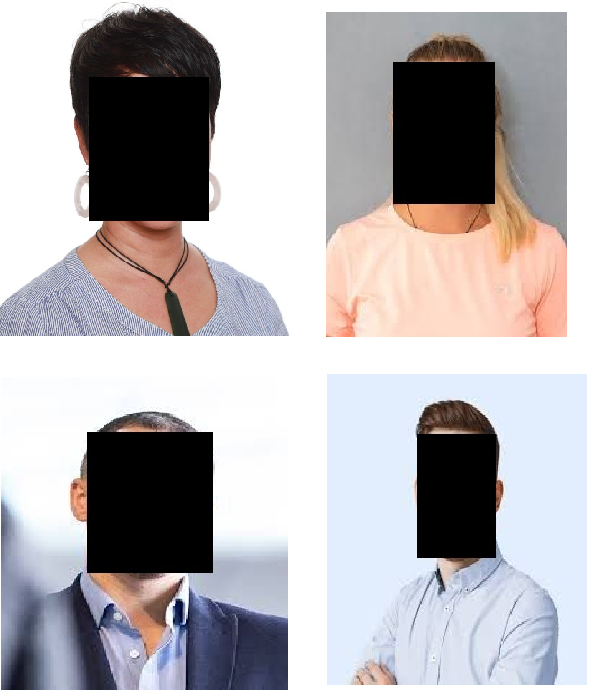}          & \includegraphics[width=0.2\textwidth]{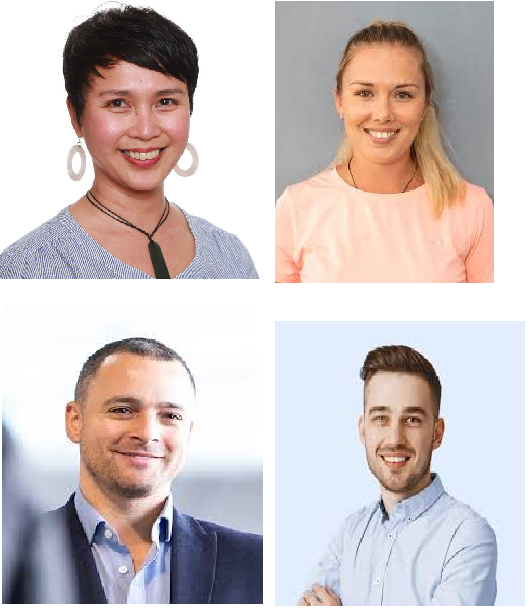}            \\ \hline
\end{tabular}
\caption{Target images}
\label{tab:table_1}
\end{table}

\section{Findings and Discussions}
\begin{table}[]
\centering
\fontsize{9}{8}\selectfont
\begin{tabular}{|l|l|r|l|r|l|}
\hline
\textbf{Model Type} & \textbf{Model Name} & \multicolumn{1}{l|}{\textbf{\begin{tabular}[c]{@{}l@{}}Mean \\ $\Delta$\end{tabular}}} & \textbf{\begin{tabular}[c]{@{}l@{}}Average \\ Model $\Delta$\end{tabular}} & \multicolumn{1}{l|}{\textbf{\begin{tabular}[c]{@{}l@{}}Mean \\  \% $\Delta$\end{tabular}}} & \textbf{\begin{tabular}[c]{@{}l@{}}Average \\ Model \%$\Delta$\end{tabular}} \\ \hline
\textbf{CNN}        & Inception           & 0.1                                                                                 & 0.11                                                                    & 15                                                                                        & 16.88                                                                     \\
                    & ResNet152           & 0.18                                                                                &                                                                         & 24.24                                                                                     &                                                                           \\
                    & VGG16               & 0.1                                                                                 &                                                                         & 18.36                                                                                     &                                                                           \\
                    & Xception            & 0.06                                                                                &                                                                         & 10                                                                                        &                                                                           \\ \hline
\textbf{ViT}        & ViT-B16             & 0.17                                                                                & \begin{tabular}[c]{@{}l@{}}0.17  \textcolor{red}{(54\% $\uparrow$)}\end{tabular}           & 39.19                                                                                     & \begin{tabular}[c]{@{}l@{}}37.8  \textcolor{red}{(123\% $\uparrow$)}\end{tabular}             \\
                    & ViT-B32             & 0.18                                                                                &                                                                         & 39                                                                                        &                                                                           \\
                    & ViT-L16             & 0.13                                                                                &                                                                         & 31                                                                                        &                                                                           \\
                    & ViT-L32             & 0.2                                                                                 &                                                                         & 42                                                                                        &                                                                           \\ \hline
\end{tabular}
\caption{Accuracy Difference ($\Delta$) for CNNs and ViTs. ($\uparrow$) indicates higher bias in percentage and is given in \textcolor{red}{red}. \%  $\Delta = \frac{\lvert A_{unbiased}-A_{biased}\lvert}{A_{unbiased}}*100$}
\label{tab:table_2}
\end{table}

\begin{table}[]
\fontsize{7}{7}\selectfont
\begin{tabular}{|l|rrrr|rrrr|}
\hline
 & \multicolumn{1}{l}{\textbf{}} & \multicolumn{1}{l}{\textbf{Masked}} & \multicolumn{1}{l}{} & \multicolumn{1}{l|}{} & \multicolumn{1}{l}{\textbf{}} & \multicolumn{1}{l}{\textbf{Unmasked}} & \multicolumn{1}{l}{} & \multicolumn{1}{l|}{} \\ \cline{2-9} 
 & \multicolumn{1}{l}{\textbf{Biased}} & \multicolumn{1}{l|}{} & \multicolumn{1}{l}{\textbf{Unbiased}} & \multicolumn{1}{l|}{} & \multicolumn{1}{l}{\textbf{Biased}} & \multicolumn{1}{l|}{} & \multicolumn{1}{l}{\textbf{Unbiased}} & \multicolumn{1}{l|}{} \\ \hline
\textbf{Class} & \multicolumn{1}{l|}{\textbf{CNN}} & \multicolumn{1}{l|}{\textbf{ViT}} & \multicolumn{1}{l|}{\textbf{CNN}} & \multicolumn{1}{l|}{\textbf{ViT}} & \multicolumn{1}{l|}{\textbf{CNN}} & \multicolumn{1}{l|}{\textbf{ViT}} & \multicolumn{1}{l|}{\textbf{CNN}} & \multicolumn{1}{l|}{\textbf{ViT}} \\ \hline
\textbf{CEO} & \multicolumn{1}{r|}{0.059} & \multicolumn{1}{r|}{0.1} & \multicolumn{1}{r|}{0.26} & 0.02 & \multicolumn{1}{r|}{0.05} & \multicolumn{1}{r|}{0.17} & \multicolumn{1}{r|}{0.07} & 0.06 \\
\textbf{Engineer} & \multicolumn{1}{r|}{0.23} & \multicolumn{1}{r|}{0.14} & \multicolumn{1}{r|}{0.36} & 0.17 & \multicolumn{1}{r|}{0.18} & \multicolumn{1}{r|}{0.19} & \multicolumn{1}{r|}{0.04} & 0.21 \\
\textbf{Nurse} & \multicolumn{1}{r|}{-0.14} & \multicolumn{1}{r|}{-0.35} & \multicolumn{1}{r|}{-0.05} & -0.2 & \multicolumn{1}{r|}{-0.21} & \multicolumn{1}{r|}{-0.21} & \multicolumn{1}{r|}{-0.06} & -0.17 \\
\textbf{School Teacher} & \multicolumn{1}{r|}{-0.17} & \multicolumn{1}{r|}{-0.15} & \multicolumn{1}{r|}{-0.12} & -0.05 & \multicolumn{1}{r|}{-0.02} & \multicolumn{1}{r|}{-0.4} & \multicolumn{1}{r|}{-0.04} & -0.14 \\ \hline
\textbf{Total IIAS (absolute)} & \multicolumn{1}{r|}{0.599} & \multicolumn{1}{r|}{0.74} & \multicolumn{1}{r|}{0.79} & 0.44 & \multicolumn{1}{r|}{0.46} & \multicolumn{1}{r|}{0.97} & \multicolumn{1}{r|}{0.21} & 0.58 \\
\textbf{\% Difference} & \multicolumn{1}{l|}{} & \multicolumn{1}{l|}{\textcolor{red}{23\% $\uparrow$}} & \multicolumn{1}{l|}{\textcolor{red}{80\% $\uparrow$}} & \multicolumn{1}{l|}{} & \multicolumn{1}{l|}{} & \multicolumn{1}{l|}{\textcolor{red}{111\% $\uparrow$}} & \multicolumn{1}{l|}{} & \multicolumn{1}{l|}{\textcolor{red}{176\% $\uparrow$}} \\ \hline
\end{tabular}
\caption{Image-Image Association Score for CNNs and ViTs. The values are the average of all the models averaged over five iterations. A +ve value indicates a bias towards men and a -ve value indicates a bias towards women. The total IIAS is calculated by adding the absolute values of the individual IIAS scores which capture bias magnitude. This is done to provide a better comparison between the models. ($\uparrow$) indicates higher IIAS i.e. higher bias in percentage and is given in \textcolor{red}{red}.}
\label{tab:table_3}
\end{table}

\begin{table}[]
\centering
\fontsize{8}{8}\selectfont
\begin{tabular}{lllll}
\hline
\textbf{Image Encoder} & \textbf{\begin{tabular}[c]{@{}l@{}}Man\\ Occurrence\end{tabular}} & \textbf{\begin{tabular}[c]{@{}l@{}}Top 3\\ Predictions\end{tabular}} & \textbf{\begin{tabular}[c]{@{}l@{}}Woman\\ Occurrence\end{tabular}} & \textbf{\begin{tabular}[c]{@{}l@{}}Top 3\\ Predictions\end{tabular}} \\ \hline
RN 50 & 47 & \begin{tabular}[c]{@{}l@{}}mathematician,\\ psychiatrist'youtuber\end{tabular} & 49 & \begin{tabular}[c]{@{}l@{}}beautician,\\ student,\\ housekeeper\end{tabular} \\ \hline
RN 50x4 & 46 & \begin{tabular}[c]{@{}l@{}}investment banker,\\ economist,\\ coach\end{tabular} & 56 & \begin{tabular}[c]{@{}l@{}}housekeeper,\\ jewellery maker,\\ midwife\end{tabular} \\ \hline
ViT B/16 & 50 & \begin{tabular}[c]{@{}l@{}}coach,\\ psychiatrist,\\ administrator\end{tabular} & 54 & \begin{tabular}[c]{@{}l@{}}midwife,\\ beautician,\\ jewellery maker\end{tabular} \\ \hline
ViT B/32 & 45 & \begin{tabular}[c]{@{}l@{}}chief executive officer,\\ musician,\\ hairdresser\end{tabular} & 63 & \begin{tabular}[c]{@{}l@{}}beautician,\\ housekeeper,\\ jewellery maker\end{tabular} \\ \hline
\textbf{CNN} & 46.5 &  & 52.5 &  \\
\textbf{ViT} & 48 \textcolor{red}{(3.3 \% $\uparrow$)} &  & 59 \textcolor{red}{(12.53 \% $\uparrow$)} &  \\ \hline
\end{tabular}
\caption{Top 3 predictions for images of men and women using CLIP. The occurrence values show the percentage of predictions for the top 3 predictions. ($\uparrow$) indicates a higher concentration of biased predictions i.e. higher bias in percentage and is given in \textcolor{red}{red}.}
\label{tab:table_4}
\end{table}

\begin{table}[]
\centering
\fontsize{8}{8}\selectfont
\begin{tabular}{|ll|l|l|}
\hline
\multicolumn{1}{|l|}{\textbf{Encoder Type}} & \begin{tabular}[c]{@{}l@{}}\textbf{Image}\\ \textbf{Encoder}\end{tabular} & \textbf{Skewness} &  \\ \hline
 &  & \textbf{Man} & \textbf{Woman} \\ \hline
\multicolumn{1}{|l|}{CNN} & RN 50 & 2.27 & 3.6 \\ \cline{2-4} 
\multicolumn{1}{|l|}{} & RN 50x4 & 2.06 & 3.84 \\ \hline
\multicolumn{1}{|l|}{ViT} & ViT-B/16 & 2.54 & 3.75 \\ \cline{2-4} 
\multicolumn{1}{|l|}{} & ViT-B/32 & 2.73 & 4.26 \\ \hline
\multicolumn{1}{|l|}{Model Average} & CNN & 2.16 & 3.7 \\ \cline{2-4} 
\multicolumn{1}{|l|}{} & ViT & 2.63 \textcolor{red}{(21.7\% $\uparrow$)} & 4 \textcolor{red}{(8 \% $\uparrow$)} \\ \hline
\end{tabular}
\caption{Skewness in CLIP's predictions using different image encoders. ($\uparrow$) indicates a higher skewness of biased predictions i.e. higher bias in percentage and is given in \textcolor{red}{red}.}
\label{tab:table_5}
\end{table}
\subsection{Accuracy Difference}
We found the Accuracy Difference for ViTs to be significantly higher than CNNs. The figures in Table~\ref{tab:table_2} show $\Delta$ to be 54\% higher and the \% $\Delta$ to be 123\% higher for ViTs. This means the effect of gender bias is higher on the ViTs. This may be explained by the fact that ViTs have global attention which enables them to  get more visual cues allowing them to deduce gender from multiple visual features. We also see the variation in $\Delta$ among the CNNs. ResNet 152 has the highest $\Delta$ and \%  $\Delta$. This may be due to ResNet 152 having a larger receptive field~\cite{distillComputingReceptive} enabling it to gather more visual information related to gender. The differences among ViTs, though not as prominent as CNNs, still show some variation with models having a larger patch size (ViT-B/32 and L/32) having more bias. As larger patch sizes enable the capture of more global information~\cite{kolesnikov2021image, parkvision, khan2022transformers}, the model can learn more information related to gender, thereby contributing to bias, in a way similar to the CNNs. 

\subsection{IIAS}
The results of the IIAS experiment showed similar results to those in the previous experiment with ViTs showing higher bias than CNNs as shown in Table~\ref{tab:table_3}. The scores show stereotypical bias in occupations with `CEO' and `Engineer' having a positive score indicating male bias and `Nurse' and `School Teacher' showing female bias as indicated by a negative score. This is similar to the results shown in previous research~\cite{mandal2023multimodal}. For the masked images, we see a 23\% higher IIAS for the biased ViT models but an 80\% higher IIAS for the unbiased CNN models.  In the case of the unmasked images, the ViTs had a higher IIAS for both the biased and unbiased models, 111\% and 176\% respectively. Ideally, as there is an equal number of images of men and women in the target sets, the values should be zero or very close. In the case of masked images, where the face is hidden, the models may learn gender from other features such as the dress worn~\cite{wang2019balanced}. ViTs with their global attention may amplify bias due to this as seen from Table~\ref{tab:table_3}. An interesting observation is that for masked images, the unbiased CNNs show a higher bias than the ViTs. This may be due to convolutions being a high-pass filter amplifying high-frequency signals~\cite{parkvision} and the absence of  the low-frequency signals in the face affecting its performance. Another reason may simply be that the CNNs are unable to localize their focus as faces generally have a higher saliency. We are, however, not fully sure of what might cause this.
\subsection{Analysis of CLIP Zero-shot Predictions}
The predictions using CLIP zero-shot (Table \ref{tab:table_4}) reveal the presence of gender bias in the model with the top three predictions for men being stereotypically male-dominated occupations such as `chief executive officer, `economist', and `investment banker' whereas those for women are stereotypically female-dominated such as `beautician', `housekeeper', and `jewellery maker'~\cite{mandal2023multimodal,wang2022revise}. The predictions are highly skewed with these biased predictions making up nearly half of all the predictions. The skewness is higher when ViTs are used as image encoders showing a higher bias. The skewness metrics given in Table~\ref{tab:table_5} also show higher skewness for ViT encoders. Although the higher bias in CLIP's ViT encoder models shows a similar pattern to our classifier experiments, the effect is less pronounced. This may be due to the debiasing done in CLIP~\cite{radford2021learning}.

\section{Conclusion and Future Work}
In our experiments, we found evidence that the model architecture affects the amplification of social biases and show that vision transformers amplify gender bias more than convolutional neural networks. We attribute this to two features of vision transformers: 1) a shallower loss landscape leading to better generalisation and 2) global attention and a larger receptive field due to the multi-headed self-attention mechanism that enables vision transformers to capture more visual cues and long-term dependencies. Both these properties of vision transformers allow them to learn contextual information and generalise better than convolutional neural networks and learn complex concepts. But this inadvertently enables ViTs to learn social concepts such as gender. Therefore, when the training data is gender biased, the ViTs learn biased associations better than CNNs. 

This paper also introduces \textit{Accuracy Difference}, a metric for social bias in both CNNs and ViTs. It may be used for estimating and comparing bias in many different types of models with different architectures. It is simple, easy to understand
and implement and can work on black box models such as closed-sourced models and APIs. We further adapted the \textit{Image-Image Association Score} for detecting bias in image classifiers and evaluated the effect of architecture choice in image encoders of a large multimodal model, CLIP. With the prevalence of large multimodal models and their wide applications, the potential for inadvertent amplification of biases is of particular concern and requires further consideration beyond gender in a binary sense and also to include other forms of social bias (geographic, racial, etc).  

\subsection{Future Work}
This research can help understand the effect of model architecture on social biases and assist developers in making informed choices about selecting vision models. One such case is CLIP, as discussed earlier. Accuracy difference can be used for bias analytics for different architectures. ViTs have been shown to outperform CNNs in many applications \cite{khan2022transformers,kolesnikov2021image,naseer2021intriguing}, leading to widespread adoption. However, if, as this research suggests, they may amplify bias to a greater extent, this aspect needs to be understood and considered as part of the adoption of ViTs.

\section{Acknowledgements}
Abhishek Mandal was partially supported by the $<$A+$>$ Alliance / Women at the Table as an Inaugural Tech Fellow 2020/2021. This publication has emanated from research supported by Science Foundation Ireland (SFI) under Grant Number SFI/12/RC/2289\textunderscore2, cofunded by the European Regional Development Fund.

We would like to thank Dr. Alessandra Mileo for her input in this paper.
\section{Appendix A}

\textbf{List of Occupations}

accountant, administrator, architect, artist, athlete, attendant, auctioneer, author, baker, beautician, blacksmith, broker, business analyst, carpenter, cashier, chef, chemist, chief executive officer, cleaner, clergy, clerk, coach, collector, conductor, construction worker, counsellor, customer service executive, dancer, dentist, designer, digital content creator, doctor, driver, economist, electrician, engineer, farmer, filmmaker, firefighter, fitter, food server, gardener, geologist, guard, hairdresser, handyman, housekeeper, inspector, instructor, investment banker, jewellery maker, journalist, judge, laborer, lawyer, librarian, lifeguard, machine operator, manager, mathematician, mechanic, midwife, musician, nurse, official, operator, painter, photographer, physician, physicist, pilot, plumber, police, porter, postmaster, product owner, professor, programmer, psychiatrist, psychologist, retail assistant, sailor, salesperson, scientist, secretary, sheriff, soldier, statistician, student, supervisor, supply chain associate, support worker, surgeon, surveyor, tailor, teacher, trainer, warehouse operative, welder, youtuber

\textbf{Sources:}~\citet{garg2018word}, BBC Careers \footnote{\url{https://www.bbc.co.uk/bitesize/articles/zdqnxyc}}, LinkedIn \footnote{\url{https://business.linkedin.com/talent-solutions/resources/talent-acquisition/jobs-on-the-rise-nl-en-cont-fact} accessed: 19-04-2023} \footnote{\url{https://business.linkedin.com/content/dam/me/business/en-us/talent-solutions/emerging-jobs-report/Emerging_Jobs_Report_U.S._FINAL.pdf} accessed: 19-04-2023}, Australian Occupation List \footnote{\url{https://immi.homeaffairs.gov.au/visas/working-in-australia/skill-occupation-list} accessed: 19-04-2023} and Canadian Occupation List\footnote{\url{https://www.canada.ca/en/immigration-refugees-citizenship/services/immigrate-canada/express-entry/eligibility/find-national-occupation-code.html} accessed: 19-04-2023}.

\bibliography{egbib}
\end{document}